\documentclass[letterpaper, 10 pt, conference]{ieeeconf}  

\IEEEoverridecommandlockouts                              

\usepackage{graphicx}
\usepackage{amsmath,amssymb,amsfonts}
\usepackage{array}
\usepackage{booktabs}
\usepackage{makecell}
\usepackage{multirow}
\usepackage{xcolor}
\usepackage{soul}
\usepackage{hyperref}
\usepackage[normalem]{ulem}
\usepackage[font=footnotesize,labelformat=simple]{subcaption}

\soulregister{\cite}{7}
\sethlcolor{yellow}

\title{\LARGE \bf
Look Forward to Walk Backward: Efficient Terrain Memory for Backward Locomotion with Forward Vision
}

\author{Shixin Luo$^{1}$, Songbo Li$^{1}$, Yuan Hao$^{1}$, Yaqi Wang$^{2}$, Jun Zheng$^{3}$, Jun Wu$^{1}$, and Qiuguo Zhu$^{*1}$ 
\thanks{This work was supported by the National Key R\&D Program of China (Grant No. 2022YFB4701502), the ”Leading Goose” R\&D Program of Zhejiang (Grant No. 2023C01177), and the 2035 Key Technological Innovation Program of Ningbo City (Grant No. 2024Z300).}
\thanks{$^{1}$The authors are with Institute of Cyber-Systems and Control, Zhejiang University, 310027, China.}%
\thanks{$^{2}$The author is with College of Artificial Intelligence, Nankai University, 300350, China.}%
\thanks{$^{3}$The author is with Hangzhou Public Library, 310016, China.}%
\thanks{$^*$Qiuguo Zhu ({\tt\small qgzhu@zju.edu.cn}) is the corresponding author.}%
\thanks{The supplementary video is available at \url{https://youtu.be/qWLktK1teIo}.}
}

\begin{document}

\maketitle
\thispagestyle{empty}
\pagestyle{empty}

\begin{abstract}
Legged robots with egocentric forward-facing depth cameras can couple exteroception and proprioception to achieve robust forward agility on complex terrain. When these robots walk backward, the forward-only field of view provides no preview. Purely proprioceptive controllers can remain stable on moderate ground when moving backward but cannot fully exploit the robot's capabilities on complex terrain and must collide with obstacles. We present Look Forward to Walk Backward (LF2WB), an efficient terrain-memory locomotion framework that uses forward egocentric depth and proprioception to write a compact associative memory during forward motion and to retrieve it for collision-free backward locomotion without rearward vision. The memory backbone employs a delta-rule selective update that softly removes then writes the memory state along the active subspace. Training uses hardware-efficient parallel computation, and deployment runs recurrent, constant-time per-step inference with a constant-size state, making the approach suitable for onboard processors on low-cost robots. Experiments in both simulations and real-world scenarios demonstrate the effectiveness of our method, improving backward agility across complex terrains under limited sensing.
\end{abstract}

\section{INTRODUCTION}

Legged robots with egocentric forward-facing depth cameras can tightly couple exteroception and proprioception to realize robust forward agility, including jumping, climbing, and negotiating tight spaces~\cite{zhuang2023robot,cheng2023extreme,chane2024soloparkour,lai2025world}. When the robot must walk backward, the forward-only field of view provides no preview. Purely proprioceptive controllers can remain stable on moderate terrain, but they follow a feel-first, adapt-later pattern that often collides with obstacles and underperforms in complex scenes where anticipating contact is essential~\cite{li2025move}. This raises a concrete question: after traversing a scene forward once, can the robot equipped only with a forward-facing camera reuse a compact terrain memory to anticipate obstacles and achieve collision-free backward locomotion without any rearward exteroception?

One seemingly straightforward option is to insert explicit hierarchical mapping, for example elevation or voxel maps, combined with SLAM for long-term memory for backward locomotion~\cite{wang2024sf}. However, in highly dynamic locomotion, such approaches add latency, accumulate drift, and propagate integration noise. This motivates a single-stage training paradigm that learns to store and retrieve terrain memory directly within the control loop, reduces the representational mismatch between modules, and uses a lightweight backbone that fits onboard processors on low-cost robots. Prior work has shown the potential of one-stage pipelines incorporating an asymmetric actor–critic and a LSTM/GRU-based estimator in blind and perceptive locomotion tasks~\cite{luo2024pie,li2025move,11155164}.

\begin{figure}[t]
\centering
\includegraphics[width=\columnwidth]{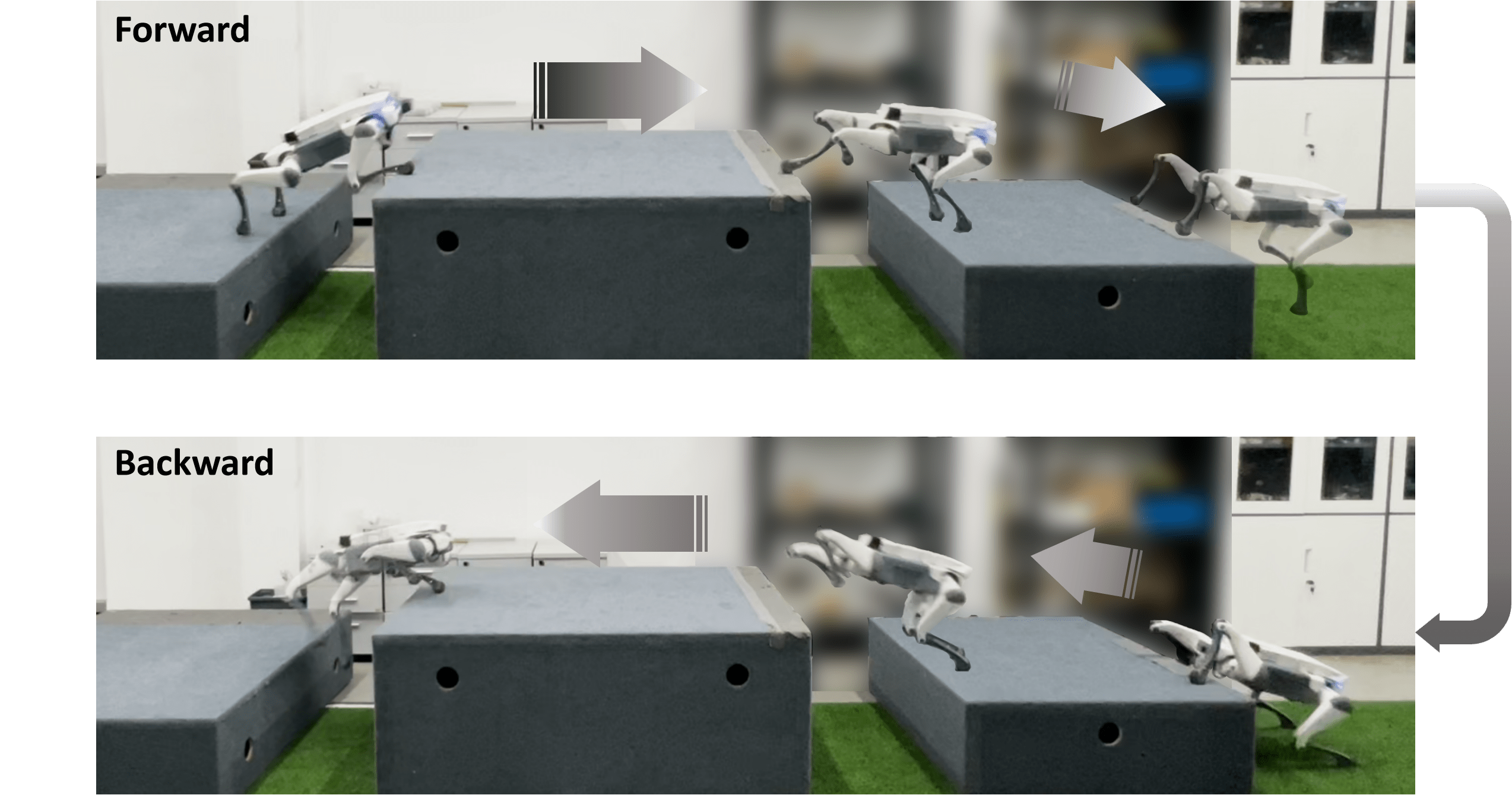}
\vspace{-0.5cm}
\captionsetup{font=footnotesize}
\caption{We deploy our policy in real-world environments, demonstrating agile forward and backward motion over discrete terrains that combine steps and gaps. The robot memorizes the surrounding terrain during forward motion and retrieves useful information from the memory state when moving backward, enabling highly dynamic backward locomotion using only a forward-facing depth camera.}
\label{fig_1}
\vspace{-0.5cm}
\end{figure}

Classical recurrent units such as LSTM and GRU have therefore been popular in resource-constrained robot learning. They offer constant per-step inference time with respect to context length, a fixed-size recurrent state, and the ability to span long episodes by passing the fixed-size memory hidden state between rollouts during on-policy data collection while truncating gradients at rollout boundaries for Truncated Backpropagation Through Time (TBPTT)~\cite{williams1990efficient} to reduce computational cost and memory usage. However, stacked LSTMs/GRUs often suffer from vanishing or exploding gradients, weak long-range modeling, and poor scaling at large batch or sequence lengths, which limits their utility when long and cluttered traversals require high-capacity memory~\cite{pascanu2013difficulty,vaswani2017attention}.

By contrast, decoder-only self-attention Transformers excel at long-range dependencies and scale incredibly well~\cite{kaplan2020scaling}. Several works on locomotion have adopted Transformer backbones, but the effective memory horizons remain too short to support long-distance backward locomotion that relies on stored terrain~\cite{radosavovic2023learning,radosavovic2024real,radosavovic2024learning}. The bottleneck is practical: training with backbones like Transformer-XL requires storing KV states to preserve memory from previous rollout segments~\cite{dai2019transformer,zhenghuwo}, so GPU memory grows with the desired context length and forces short memory windows~\cite{lin2025let}. This is a disadvantage relative to RNNs, which carry only a constant-size recurrent state. At inference time, attention with KV caching incurs per-step cost that increases with context length since the model attends over all cached keys, which conflicts with the need for stable, constant-time control cycles on real robots and again encourages aggressive memory context window clipping. Moreover, because memory is represented per time step, every step appends KV entries. In our task, if the robot waits in place, the local terrain map should remain unchanged, but step-wise caching keeps growing with redundant entries until the cache limit is reached, at which point earlier terrain is forgotten simply due to limited KV cache capacity.

Modern recurrent backbones have recently closed much of the gap between classic gated RNNs and self-attention Transformers~\cite{katharopoulos2020transformers,sun2023retentive,yang2023gated,gu2023mamba}. Unlike stacked LSTMs/GRUs, they train efficiently with parallel or chunkwise parallel algorithms and exhibit better stability and scaling at longer horizons. Unlike decoder-only self-attention Transformers, they require no KV cache, so inference time per step and memory footprint remain essentially constant with respect to context length, which keeps control cycles stable for onboard processors. These backbones also strengthen long-range modeling through data-aware state updates and structured transitions, yielding compact, durable associative memory that is particularly suitable for storing and retrieving traversed terrain during backward locomotion.

From the perspective of memory mechanisms developed for language models, our setting can be viewed as an in-context retrieval task: forward motion must write terrain cues that remain retrievable much later when the robot reverses. Plain linear attention is not suitable because it only accumulates key–value associations and lacks a principled removal mechanism, which leads to overload and degraded retrieval as sequences grow. A common fix is to add an explicit forget gate. In language models, a forget gate is valuable for rapidly clearing outdated or irrelevant information during context switches. In backward locomotion with forward-only vision, however, the memory should behave more like an implicit SLAM store, where information encountered long ago can become critical when the robot moves back, and aggressive decay risks erasing exactly that information. Training with rollout windows further amplifies this effect because state is detached across windows, so gradient does not propagate globally and explicit forgetting accelerates erosion of distant but relevant terrain memory. Among modern recurrent models, delta-rule variants such as DeltaNet perform especially well on in-context retrieval because each step first removes outdated content tied to the current input and then adds a calibrated update from the new observation~\cite{widrow1988adaptive,schlag2021linear,yang2024parallelizing}. This selective overwrite avoids uncontrolled accumulation without invoking rapid global forgetting.

Inspired by these works, we propose Look Forward to Walk Backward (LF2WB), a memory-enhanced perceptive locomotion controller that encodes forward egocentric depth and proprioception into a compact, associative terrain memory using a delta-rule selective-update backbone DeltaNet. The policy conditions backward actions on retrieved memory and proprioception to anticipate obstacles without contact. During training we carry only a constant-size recurrent state across rollout windows and use chunkwise parallel updates inside each window. At deployment, LF2WB preserves terrain memory over extended horizons and executes agile backward traversals over steps and gaps as shown in Fig.~\ref{fig_1}. When the robot needs to retreat after forward traversal, it does not turn in place. It simply reverses the previous action sequence, and as long as the terrain was traversable in the forward direction, our memory framework provides accurate backward terrain information to guide the return.

\begin{figure*}[t]
\centering
\vspace{0.2cm}
\includegraphics[width=\textwidth]{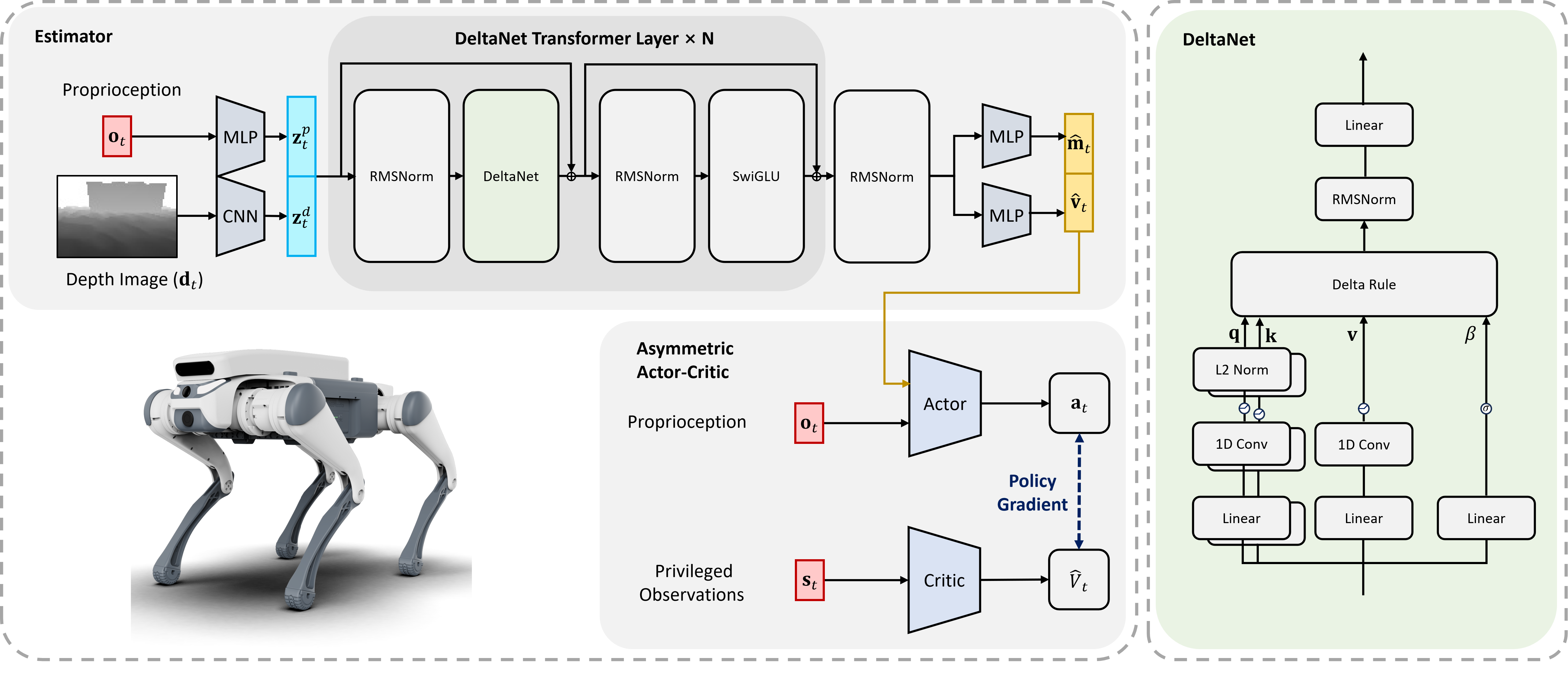}
\vspace{-0.5cm}
\captionsetup{font=footnotesize}
\caption{Overview of the proposed LF2WB framework. We concurrently train a delta-rule estimator and an asymmetric actor-critic. The estimator learns a compact terrain memory with strong retrieval ability for backward motion, enabling the policy to execute dynamic backward maneuvers using only forward egocentric vision.}
\vspace{-0.5cm}
\label{fig_2}
\end{figure*}

Our contributions are as follows:
\begin{itemize}
\item{We present Look Forward to Walk Backward (LF2WB), a single-stage training framework that learns to write terrain memory during forward motion and read it for collision-free backward locomotion without rearward vision.}
\item{We integrate a delta-rule update mechanism that retains implicit terrain information, preventing uncontrolled memory accumulation while avoiding aggressive memory resets. This structure uses a constant-size recurrent state, supports chunkwise parallel training, and offers constant-time, constant-memory per-step inference with no KV cache, which makes it practical for low-cost onboard processors.}
\item{We demonstrate zero-shot real-world deployment on a low-cost quadruped robot, achieving highly dynamic agile backward maneuvers over challenging step and gap terrains using only forward egocentric vision.}
\end{itemize}

\section{RELATED WORK}
\label{chap:2}

\subsection{Learning-based Legged Locomotion}

Recent learning-based blind controllers leverage reliable proprioception and achieve agile, robust motion in simulation and on hardware. They learn quickly, support online fine tuning, and withstand strong disturbances~\cite{hwangbo2019learning,lee2020learning,fu2021minimizing,rudin2022learning,smith2022legged,margolis2022rapid,margolis2023walk}.
Other works enhance blind control with adaptation from privileged signals, or by training the policy together with a state estimator or implicit imagination modules~\cite{kumar2021rma,yu2020learning,ji2022concurrent,nahrendra2023dreamwaq,long2023hybrid,wang2024cts,luo2024moral}. Map-centric perceptive pipelines use elevation, traversability, or voxel representations to perceive terrain and have shown promising results. They often require well-tuned and heavy upstream modules to avoid latency, drift, and integration noise on fast or highly dynamic motions~\cite{fankhauser2018probabilistic,pan2019gpu,miki2022elevation,hoeller2022neural,miki2022learning,hoeller2024anymal,wang2024sf,he2025attention}.
Direct cross-modal fusion of forward vision with proprioception enables vigorous and agile motion on complex terrain~\cite{loquercio2023learning,fu2022coupling,kareer2023vinl,wang2025more}. Adding memory modules, such as transformers or LSTM/RNNs, improves the use of historical information. This helps state estimation in partially observable settings and supports remembering terrain that is not currently visible underfoot~\cite{zhuang2023robot,cheng2023extreme,agarwal2023legged,radosavovic2023learning,radosavovic2024real,radosavovic2024learning,luo2024pie,li2025move,lai2025world,yang2025agile,lin2025let,zhenghuwo,11155164}.

\subsection{Linear-time Recurrent Sequence Models}

Early linear-time formulations showed that causal attention can be rewritten as kernelized associative scans with recurrent state updates, reducing complexity to $O(L)$ for a sequence of length $L$ while preserving a parallel training view~\cite{katharopoulos2020transformers}. This perspective links naturally to fast-weight memory: linearized attention programs a finite associative memory via outer-product updates, and delta-rule corrections address capacity and editability issues in purely additive schemes~\cite{schlag2021linear}. Building on this fast-weight view, delta-rule linear RNNs developed a remove–then–write update together with a Householder/WY-based sequence-parallel training algorithm that scales to long sequences while maintaining linear-time inference~\cite{yang2024parallelizing}. In parallel, retentive networks instantiated a retention operator with a constant-size recurrent state at inference and explicitly described parallel, recurrent, and chunkwise-recurrent computation forms~\cite{sun2023retentive}. Hybrid recurrent–Transformer designs pursued linear-time decoding at scale~\cite{peng2023rwkv}. Structured state-space modeling introduced selective, input-dependent dynamics with hardware-aware selective-scan kernels for linear-time inference~\cite{gu2023mamba}, and a subsequent unifying framework connected attention and SSMs via semiseparable operators with improved algorithms~\cite{dao2024transformers}. Within the linear-attention family, hardware-efficient gated variants provided input-conditioned gating and I/O-aware kernels under chunkwise-parallel schedules~\cite{yang2023gated}, and gated-delta formulations extended delta-rule models with data-dependent decay and parallel training refinements~\cite{yang2024gated}.

\section{METHOD}
\label{chap:3}

\subsection{Overview}

We maintain an implicit internal memory constructed from forward egocentric sensing and proprioception, and use it throughout motion to support backward locomotion when rearward exteroception is unavailable. As shown in Fig.~\ref{fig_2}, the system contains three modules trained concurrently on the same on-policy rollouts:

\subsubsection{Estimator}

It fuses a forward depth image \(\mathbf{d}_t\!\in\!\mathbb{R}^{H_d\times W_d}\) and proprioception \(\mathbf{o}_t\!\in\!\mathbb{R}^{D_o}\) to predict a body-centric elevation map \(\hat{\mathbf{m}}_t\!\in\!\mathbb{R}^{H_m\times W_m}\) covering \(1.1\,\mathrm{m}\times2.5\,\mathrm{m}\), and a base linear velocity estimate \(\hat{\mathbf{v}}_t\!\in\!\mathbb{R}^{D_v}\).

\subsubsection{Actor} 
\(\pi_\theta\): takes deployable observations \([\hat{\mathbf{m}}_t,\hat{\mathbf{v}}_t,\mathbf{o}_t]\) and outputs an action \(\mathbf{a}_t\).

\subsubsection{Critic} \(V_\phi\): takes privileged inputs \([\mathbf{m}_t,\mathbf{v}_t,\mathbf{o}_t]\) and predicts value \(V_t\).

\subsection{Estimator}

\subsubsection{Encoders and Cross-Modal Fusion}
Depth and proprioception are encoded separately, then fused:
\begin{align}
\mathbf{z}^{(d)}_t &= f_{\mathrm{cnn}}(\mathbf{d}_t) \in \mathbb{R}^{D_d},\\
\mathbf{z}^{(p)}_t &= f_{\mathrm{mlp}}(\mathbf{o}_t) \in \mathbb{R}^{D_p},\\
\mathbf{x}_t &= [\,\mathbf{z}^{(d)}_t;\,\mathbf{z}^{(p)}_t\,] \in \mathbb{R}^{D_x}.
\end{align}
The fused token \(\mathbf{x}_t\) feeds a stack of \(L\) DeltaNet-Transformer layers.

\subsubsection{DeltaNet-Transformer Layer}
We follow a LLaMA-style block and replace self-attention with a DeltaNet layer~\cite{yang2024parallelizing}. Each layer uses RMSNorm pre-norm, applies normalization before the output projection, and is followed by a SwiGLU feed-forward network.

Let the layer input be \(\mathbf{x}^{(\ell-1)}_t\in\mathbb{R}^{d}\) and \(\tilde{\mathbf{x}}_t=\mathrm{RMSNorm}(\mathbf{x}^{(\ell-1)}_t)\).
Feature maps (SiLU + \(L_2\) normalization) define keys/queries as
\begin{align}
\mathbf{k}^{(\ell)}_t &= 
  \frac{\mathrm{SiLU}\!\big(\mathbf{W}^{(\ell)}_{K}\tilde{\mathbf{x}}_t\big)}
       {\big\|\mathrm{SiLU}\!\big(\mathbf{W}^{(\ell)}_{K}\tilde{\mathbf{x}}_t\big)\big\|_2},\\
\mathbf{q}^{(\ell)}_t &= 
  \frac{\mathrm{SiLU}\!\big(\mathbf{W}^{(\ell)}_{Q}\tilde{\mathbf{x}}_t\big)}
       {\big\|\mathrm{SiLU}\!\big(\mathbf{W}^{(\ell)}_{Q}\tilde{\mathbf{x}}_t\big)\big\|_2}.
\end{align}
The attention value vector (not the robot velocity) is
\begin{equation}
\mathbf{v}^{(\ell)}_{t,\mathrm{att}}=\mathbf{W}^{(\ell)}_{V}\tilde{\mathbf{x}}_t\in\mathbb{R}^{d_v},
\end{equation}
and the overwrite scalar is
\begin{equation}
\beta^{(\ell)}_t=\sigma\!\left({\mathbf{w}^{(\ell)}_{\beta}}^{\!\top}\tilde{\mathbf{x}}_t\right)\in(0,1).
\end{equation}
The per-layer state matrix \(\mathbf{S}^{(\ell)}_t\in\mathbb{R}^{d_v\times d_k}\) is updated by the Delta Rule (remove-then-write along \(\mathbf{k}^{(\ell)}_t\)):
\begin{align}
\mathbf{S}^{(\ell)}_t &= 
  \mathbf{S}^{(\ell)}_{t-1}\!\Big(\mathbf{I}-\beta^{(\ell)}_t\,\mathbf{k}^{(\ell)}_t{\mathbf{k}^{(\ell)}_t}^{\!\top}\Big)
  +\beta^{(\ell)}_t\,\mathbf{v}^{(\ell)}_{t,\mathrm{att}}{\mathbf{k}^{(\ell)}_t}^{\!\top},\\
\mathbf{o}^{(\ell)}_{t,\mathrm{att}} &= \mathbf{S}^{(\ell)}_t\,\mathbf{q}^{(\ell)}_t.
\end{align}
With \(L_2\)-normalized \(\mathbf{k},\mathbf{q}\), the eigenvalues of \(\mathbf{I}-\beta\,\mathbf{k}\mathbf{k}^{\!\top}\) are
\(\{1\}^{d-1}\cup\{1-\beta\}\subset[0,1]\), stabilizing training and interpreting \(\beta{=}1\) as a projection that erases one subspace while retaining the other \(d{-}1\).

Before output projection we apply RMSNorm:
\begin{align}
\hat{\mathbf{o}}^{(\ell)}_{t,\mathrm{att}} &= \mathrm{RMSNorm}\!\big(\mathbf{o}^{(\ell)}_{t,\mathrm{att}}\big),\\
\Delta\mathbf{x}^{(\ell)}_t &= \mathbf{W}^{(\ell)}_{O}\,\hat{\mathbf{o}}^{(\ell)}_{t,\mathrm{att}},\\
\mathbf{x}'^{(\ell)}_t &= \mathbf{x}^{(\ell-1)}_t + \Delta\mathbf{x}^{(\ell)}_t.
\end{align}
SwiGLU FFN with pre-norm:
\begin{align}
\bar{\mathbf{x}}^{(\ell)}_t &= \mathrm{RMSNorm}\!\big(\mathbf{x}'^{(\ell)}_t\big),\\
\mathrm{FFN}\!\big(\bar{\mathbf{x}}^{(\ell)}_t\big) 
  &= \mathbf{W}^{(\ell)}_{2}\,\mathrm{SwiGLU}\!\big(\mathbf{W}^{(\ell)}_{1}\bar{\mathbf{x}}^{(\ell)}_t\big),\\
\mathbf{x}^{(\ell)}_t &= \mathbf{x}'^{(\ell)}_t + \mathrm{FFN}\!\big(\bar{\mathbf{x}}^{(\ell)}_t\big).
\end{align}
The layer above replaces self-attention one-for-one; depth/width/FFN dimensions and residual layout match a standard Transformer block.

\subsubsection{Chunkwise Parallel Training}
For long sequences, we adopt a chunkwise parallel algorithm that vectorizes intra-chunk computation while carrying a fixed-size inter-chunk state. Split a length-\(L\) sequence into \(L/C\) chunks of size \(C\). For chunk index $u$, stack the per-token projections and overwrite gates row-wise
\begin{align}
\mathbf{Q}[u],\,\mathbf{K}[u],\,\mathbf{V}[u] &\in \mathbb{R}^{C\times d_k},\,\mathbb{R}^{C\times d_k},\,\mathbb{R}^{C\times d_v},\\
\boldsymbol{\beta}[u] &\in \mathbb{R}^{C}.
\end{align}
Using the UT transform to expose large matmuls, define
\begin{align}
\mathbf{T}[u]
  &= \Big( \mathbf{I}
     + \mathrm{strictLower}\!\big(
         \mathrm{diag}(\boldsymbol{\beta}[u])\,\mathbf{K}[u]
     \big.\nonumber\\
  &\qquad\big.
         \mathbf{K}[u]^{\!\top}
       \big) \Big)^{-1}
     \mathrm{diag}(\boldsymbol{\beta}[u]),\\
\mathbf{W}[u] 
  &= \mathbf{T}[u]\mathbf{K}[u],\\
\mathbf{U}[u] 
  &= \mathbf{T}[u]\mathbf{V}[u].
\end{align}
Let $\mathbf{S}[u]\in\mathbb{R}^{d_v\times d_k}$ be the (small) state entering chunk $u$. The cross-chunk update and the chunk outputs are
\begin{align}
\mathbf{S}[u{+}1] 
  &= \mathbf{S}[u]\,
     \Big(\mathbf{I} - \mathbf{K}[u]^{\!\top}\mathbf{W}[u]\Big)
      + \mathbf{U}[u]^{\!\top}\mathbf{K}[u],\\
\mathbf{O}[u] 
  &= \Big(\mathbf{Q}[u]\mathbf{K}[u]^{\!\top}\odot\mathbf{M}_C\Big)\,
     \Big(\mathbf{U}[u] - \mathbf{W}[u]\mathbf{S}[u]^{\!\top}\Big)
     \nonumber\\
  &\quad + \mathbf{Q}[u]\mathbf{S}[u]^{\!\top},
\end{align}
where $\mathbf{M}_C$ is the strict lower-triangular causal mask. All intra-chunk interactions are parallel and GPU-friendly; only the compact $\mathbf{S}[u]$ is propagated across chunks. No per-token hidden histories are materialized. The training complexity is $O(LCd + Ld^2)$, while test-time inference is strictly recurrent with \emph{constant} per-head state $\mathbf{S}^{(\ell)}_t$.

\subsubsection{Heads and Supervised Losses}
The top layer feature \(\mathbf{x}^{(L)}_t\) is decoded to
\begin{align}
\hat{\mathbf{m}}_t &\in \mathbb{R}^{H_m\times W_m},\\
\hat{\mathbf{v}}_t &\in \mathbb{R}^{D_v},
\end{align}
We minimize
\begin{align}
\mathcal{L}_{\mathrm{est}} 
  &= \lambda_m \big\|\hat{\mathbf{m}}_t-\mathbf{m}_t\big\|_2^2
   + \lambda_v \big\|\hat{\mathbf{v}}_t-\mathbf{v}_t\big\|_2^2.
\end{align}
where \(\lambda_m,\lambda_v>0\) are scalar weights, \(\mathbf{m}_t\) is the ground-truth elevation map, and \(\mathbf{v}_t\) is the ground-truth base linear velocity.

\begin{figure*}[t]
\centering
\includegraphics[width=\textwidth]{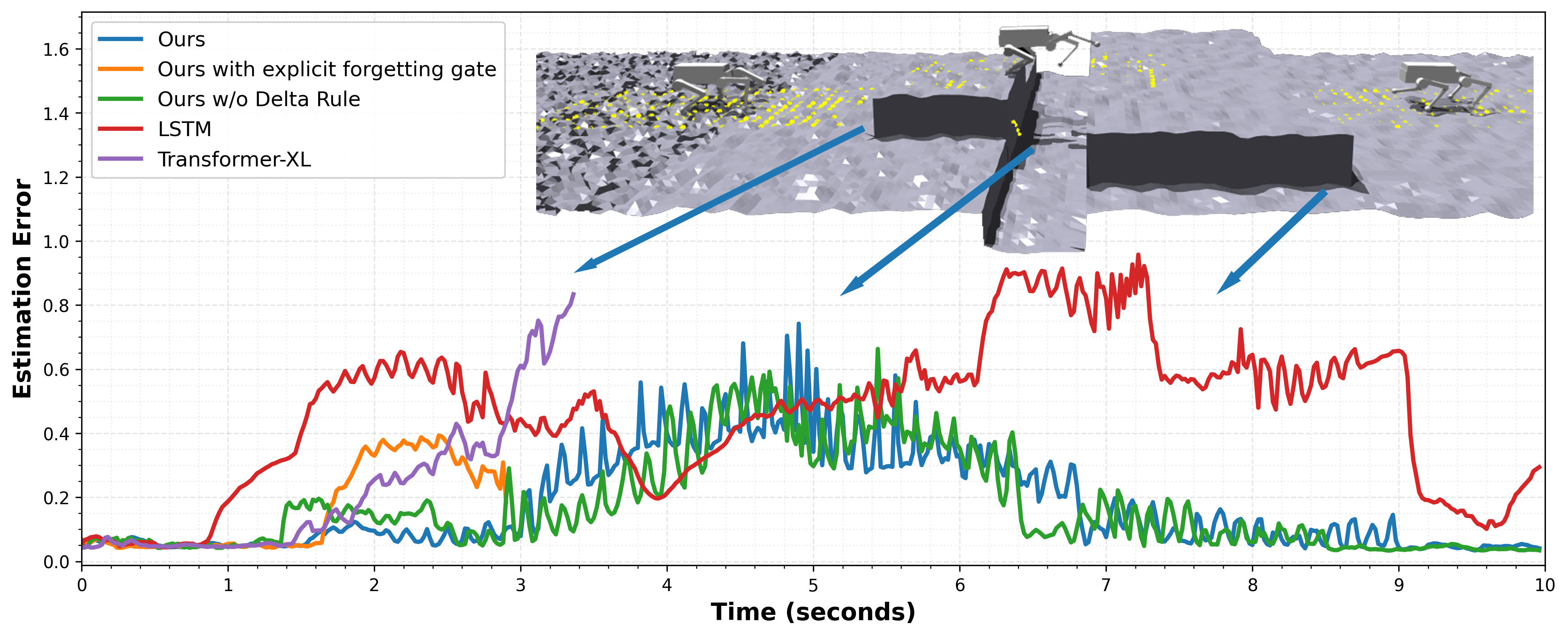}
\vspace{-0.5cm}
\captionsetup{font=footnotesize}
\caption{RMSE of the body-centric elevation estimate along the executed path. The robot starts on the left of the image and walks backward to the right; \(0\,\mathrm{s}\) corresponds to the leftmost position where the backward phase begins. Curve locations at terrain discontinuities are marked by arrows. Once the robot falls, we stop recording error.}
\label{fig_3}
\vspace{-0.5cm}
\end{figure*}

\subsection{Asymmetric Actor-Critic}

We adopt the asymmetric actor-critic framework~\cite{pinto2017asymmetric}. Only deployable signals serve as the input of the actor and privileged inputs are used to stabilize value learning:
\begin{align}
\tilde{\mathbf{s}}_t &= [\,\mathrm{vec}(\hat{\mathbf{m}}_t);\ \hat{\mathbf{v}}_t;\ \mathbf{o}_t\,],\\
\mathbf{a}_t &\sim \pi_\theta(\cdot \mid \tilde{\mathbf{s}}_t),\\
\mathbf{s}_t &= [\,\mathrm{vec}(\mathbf{m}_t);\ \mathbf{v}_t;\ \mathbf{o}_t\,],\\
V_t &= V_\phi(\mathbf{s}_t).
\end{align}
The actor is deployable; the critic uses privileged inputs for training only.

We use PPO for on-policy reinforcement learning~\cite{schulman2017proximal}. Let \(r_t(\theta)=\dfrac{\pi_\theta(\mathbf{a}_t\mid\tilde{\mathbf{s}}_t)}{\pi_{\theta_{\mathrm{old}}}(\mathbf{a}_t\mid\tilde{\mathbf{s}}_t)}\) be the importance ratio,
\(\widehat{A}_t\) the GAE advantage, and \(R_t\) the return target.
\begin{align}
r_t(\theta) &= 
  \frac{\pi_\theta(\mathbf{a}_t \mid \tilde{\mathbf{s}}_t)}
       {\pi_{\theta_{\mathrm{old}}}(\mathbf{a}_t \mid \tilde{\mathbf{s}}_t)},\\
\mathcal{L}_{\mathrm{clip}}(\theta) &= 
  \mathbb{E}\!\left[\min\!\Big(r_t \widehat{A}_t,\ 
  \mathrm{clip}(r_t,1{-}\epsilon,1{+}\epsilon)\widehat{A}_t\Big)\right],\\
\mathcal{L}_{V}(\phi) &= \mathbb{E}\!\big[(V_\phi(\mathbf{s}_t)-R_t)^2\big],\\
\mathcal{L}_{\mathrm{ent}}(\theta) &= 
  \mathbb{E}\!\big[\mathcal{H}\!\left(\pi_\theta(\cdot\mid\tilde{\mathbf{s}}_t)\right)\big],
\end{align}
where \(\mathcal{H}(\cdot)\) is entropy and \(\mathrm{clip}(\cdot)\) is applied elementwise.
The minimized PPO objective is
\begin{align}
\mathcal{L}_{\mathrm{PPO}}(\theta,\phi) &= 
  -\,\mathcal{L}_{\mathrm{clip}}(\theta)
  + c_V\,\mathcal{L}_{V}(\phi)
  - c_H\,\mathcal{L}_{\mathrm{ent}}(\theta),
\end{align}
with scalar weights \(c_V,c_H>0\).

\subsection{Training Details}
We optimize the estimator, actor, and critic concurrently, which avoids pretraining and exposes the policy to realistic estimation noise~\cite{ji2022concurrent,nahrendra2023dreamwaq}:
\begin{align}
\mathcal{L} &= \mathcal{L}_{\mathrm{PPO}} + \alpha\,\mathcal{L}_{\mathrm{est}}.
\end{align}
with \(\alpha>0\).

In the massively parallel on-policy setting, we collect short contiguous rollout windows of \(K\) steps per environment so that the update batch \(B = N_{\mathrm{env}}\!\times\!K\) remains in a regime that preserves temporal coherence while sustaining GPU throughput~\cite{rudin2022learning}. Episodes span multiple rollout updates, and we carry only the estimator’s compact recurrent state \(\mathbf{S}^{(\ell)}_t\) across windows. 

The first state of the next rollout window is detached, reducing memory. Crucially, compared to self-attention Transformers, the size of each \(\mathbf{S}^{(\ell)}_t\) is constant and does not grow with the desired memory horizon.

During rollout interaction with the simulator to collect data, the estimator runs in serial recurrent mode, updating the state \(\mathbf{S}^{(\ell)}_t\) step by step. After a window of \(K\) steps is collected, the optimization phase replays the window and computes gradients with the chunkwise parallel algorithm for the estimator.

In our setup, the maximum episode length is set to \(20\,\mathrm{s}\). Accordingly we aim for a \(20\,\mathrm{s}\) terrain memory. We set \(K{=}320\) with a \(50\,\mathrm{Hz}\) control rate, so each rollout window covers \(6.4\,\mathrm{s}\). During training we command the robot to move forward for \(10\,\mathrm{s}\) and then backward for \(10\,\mathrm{s}\), over a curriculum that progressively increases the difficulty of high platforms and gaps, as well as their random combinations. Other training configurations such as reward design, domain randomization, and PPO hyperparameters follow \cite{luo2024pie} exactly.

\section{EXPERIMENTS}
\label{chap:4}

\begin{table*}[t]
\vspace{0.3cm}
  \centering
  \captionsetup{font=footnotesize}
  \caption{Success rate (\%) across difficulty \(\delta \in \{0.2,0.4,0.6,0.8\}\) and protocol \(\textbf{P1--P5}\).}
  \label{table:1}
  \scriptsize
  \setlength{\tabcolsep}{4pt}
  \renewcommand{\arraystretch}{1.7}
  \begin{tabular}{lcccc|cccc|cccc|cccc|cccc}
  \toprule
   & \multicolumn{4}{c}{\textbf{P1 (F10\,s$\rightarrow$B10\,s)}} & \multicolumn{4}{c}{\textbf{P2 (F20\,s$\rightarrow$B10\,s)}} & \multicolumn{4}{c}{\textbf{P3 (F20\,s$\rightarrow$B20\,s)}} & \multicolumn{4}{c}{\textbf{P4 (F30\,s$\rightarrow$B10\,s)}} & \multicolumn{4}{c}{\textbf{P5 (F30\,s$\rightarrow$B30\,s)}}\\
   \cmidrule(lr){2-5}\cmidrule(lr){6-9}\cmidrule(lr){10-13}\cmidrule(lr){14-17}\cmidrule(lr){18-21}
   \textbf{Estimator Backbone} & \textbf{0.2} & \textbf{0.4} & \textbf{0.6} & \textbf{0.8} & \textbf{0.2} & \textbf{0.4} & \textbf{0.6} & \textbf{0.8} & \textbf{0.2} & \textbf{0.4} & \textbf{0.6} & \textbf{0.8} & \textbf{0.2} & \textbf{0.4} & \textbf{0.6} & \textbf{0.8} & \textbf{0.2} & \textbf{0.4} & \textbf{0.6} & \textbf{0.8}\\
  \midrule
   \makecell[l]{\vphantom{(}\textbf{Ours}} & 90.1 & 86.9 & 81.7 & 75.0 & 89.5 & 83.5 & 77.7 & 71.7 & 49.0 & 42.2 & 37.9 & 30.4 & 87.0 & 83.6 & 78.9 & 72.1 & 16.1 & 10.9 & 10.8 & 10.5 \\
   \makecell[l]{\textbf{Ours with explicit} \\ \textbf{forgetting gate}} & 70.1 & 66.4 & 57.5 & 46.7 & 72.7 & 67.4 & 59.0 & 46.6 & 25.3 & 19.1 & 10.5 & 8.8 & 66.1 & 60.8 & 56.5 & 46.5 & 9.8 & 5.1 & 3.3 & 3.4 \\
   \makecell[l]{\vphantom{(}\textbf{Ours w/o Delta Rule}} & 85.1 & 86.8 & 81.5 & 68.9 & 62.0 & 54.9 & 42.0 & 30.7 & 36.7 & 32.8 & 23.9 & 12.4 & 39.6 & 35.1 & 22.8 & 20.0 & 7.0 & 6.2 & 3.0 & 2.7 \\
   \makecell[l]{\vphantom{(}\textbf{LSTM}} & 57.1 & 45.0 & 32.8 & 25.7 & 60.3 & 48.0 & 31.3 & 25.7 & 20.6 & 7.6 & 4.3 & 3.8 & 64.3 & 52.2 & 35.8 & 29.7 & 5.8 & 2.3 & 1.0 & 1.1 \\
   \makecell[l]{\vphantom{(}\textbf{Transformer-XL}} & 28.9 & 26.6 & 22.7 & 16.7 & 30.4 & 24.7 & 20.1 & 16.2 & 7.8 & 7.7 & 3.9 & 1.9 & 28.2 & 24.7 & 19.7 & 15.5 & 2.1 & 1.2 & 1.1 & 1.0 \\
  \bottomrule
  \end{tabular}
  \vspace{-0.5cm}
\end{table*}

\subsection{Comparative Experiments}
To systematically investigate performance across complex terrains, we evaluate in Isaac Gym using the DEEP Robotics Lite3 quadruped robot and compare estimator backbones under identical overall structure, PPO settings, reward, and domain randomization; only the estimator differs. The four methods that we compared ours against are:
\begin{itemize}
\item\textbf{Ours with explicit forgetting gate (Gated DeltaNet~\cite{yang2024gated}):} 
Augments the Delta Rule update with a learned input-conditioned decay gate that down-weights the retained state before writing new content for rapid purging of spurious associations.
\item\textbf{Ours w/o Delta Rule (Linear Attention~\cite{katharopoulos2020transformers}):}
Removes the Delta Rule and keeps only additive key-value accumulation. It preserves constant-size recurrent state at inference but lacks a principled erasure mechanism.
\item\textbf{LSTM~\cite{hochreiter1997long}:}
A stacked LSTM with matched network size, trained with the same on-policy pipeline and TBPTT. It offers fixed-size recurrent state and serial inference.
\item\textbf{Transformer-XL~\cite{dai2019transformer}:}
A decoder-only Transformer that reuses cached detached hidden states from preceding segments when training.
\end{itemize}

Because Transformer-XL must retain preceding segment memories via cached hidden states whose size grows with the desired memory length and thus increases GPU memory usage, its training memory window is limited to \(240\) steps (\(4.8\,\mathrm{s}\)) versus \(320\) steps (\(6.4\,\mathrm{s}\)) for all other backbones.

To examine backward elevation estimation error, each policy controls the robot to walk forward \(10\,\mathrm{s}\) and then backward \(10\,\mathrm{s}\). As shown in Fig.~\ref{fig_3}, during the backward phase we plot the RMSE of the body-centric elevation estimate along the executed path. When the estimator's error becomes large enough that the policy can no longer negotiate the upcoming terrain and fails, the run terminates and its error trace naturally truncates at that time. From these traces, our method exhibits markedly lower error on complex terrain, supporting collision-free backward maneuvers. \textbf{Ours w/o Delta Rule} also performs well on this in-distribution \(20\,\mathrm{s}\) task, with accuracy comparable to \textbf{Ours}. \textbf{LSTM} produces less accurate estimates, yet concurrent training enables the actor to tolerate moderate bias/noise and still complete the trail. \textbf{Ours w/ explicit forgetting gate} and \textbf{Transformer-XL} show large bias and typically fail. Although self-attention is powerful, \textbf{Transformer-XL}'s effective memory is curtailed by its short segment window under our budget, preventing learning of longer-horizon terrain memory.

We then evaluate all backbones across four terrain difficulty levels \(\delta\in\{0.2,0.4,0.6,0.8\}\) with increasing complexity. Difficulty is defined by linearly scaling the maxima used during training: the hardest training terrain contains random mixtures of steps up to \(0.8\,\mathrm{m}\) in height and gaps up to \(1.1\,\mathrm{m}\) in width, combined with additional vertical offsets (e.g., a \(1.0\,\mathrm{m}\) gap whose two sides differ in height by \(0.7\,\mathrm{m}\)). A test trail at difficulty \(\delta\) uses the same random mixing but scaled linearly, i.e., \(\text{step}(\delta)=0.8\delta\,\mathrm{m}\) and \(\text{gap}(\delta)=1.1\delta\,\mathrm{m}\); for example, \(\delta{=}0.8\) yields \(0.64\,\mathrm{m}\) steps and \(0.88\,\mathrm{m}\) gaps.

We consider five forward/backward horizon protocols at each difficulty:
\begin{enumerate}
\item \textbf{P1 (F10\,s$\rightarrow$B10\,s):} \label{prot:p1}
In-distribution evaluation that matches the training schedule. This setting tests baseline backward recall.
\item \textbf{P2 (F20\,s$\rightarrow$B10\,s):} \label{prot:p2}
Capacity/overload check. An extra 10\,s of forward writing examines whether irrelevant or erroneous content can be removed without harming backward performance.
\item \textbf{P3 (F20\,s$\rightarrow$B20\,s):} \label{prot:p3}
Long-horizon recall. This setting assesses whether distant yet relevant terrain information remain retrievable when the backward horizon is doubled relative to training.
\item \textbf{P4 (F30\,s$\rightarrow$B10\,s):} \label{prot:p4}
Strong overload with a short backward retrieval. Write capacity is pushed substantially beyond training while the backward horizon remains short.
\item \textbf{P5 (F30\,s$\rightarrow$B20\,s):} \label{prot:p5}
Strong overload with a long backward retrieval. This setting stresses maximum write load together with extended backward retrieval.
\end{enumerate}

As shown in Table~\ref{table:1}, \textbf{Ours} achieves the best performance across all protocol groups and difficulty levels. Although performance decreases as the mismatch between the test horizon and the training horizon grows, our method still exhibits markedly stronger generalization than other baselines. We attribute this to the delta-rule selective update, which improves in-context recall. Training with longer memory horizons is expected to further mitigate the performance drop. Due to the explicit forgetting gate and the update scheme during training that does not optimize over the entire sequence (similar to TBPTT), \textbf{Ours with explicit forgetting gate} and \textbf{LSTM} have difficulty maintaining long-range information, leading to large drops on \textbf{P\ref{prot:p3}} and \textbf{P\ref{prot:p5}}. On the other hand, the forgetting gate can quickly remove redundant content in the memory state, so the extra forward time in \textbf{P\ref{prot:p2}} and \textbf{P\ref{prot:p4}} has limited impact early in the backward segment. \textbf{Ours w/o Delta Rule} performs well on the training-like setting but has no removal mechanism, continuously accumulating key-value outer products in a fixed-size state whose finite capacity eventually leads to overload, producing a sharper decline on \textbf{P\ref{prot:p2}} and \textbf{P\ref{prot:p4}}. Conversely, because it does not forget, it can preserve older information better than gated models on \textbf{P\ref{prot:p3}}, where long-range retrieval dominates. \textbf{Transformer-XL} underperforms due to the limited memory segment window enforced during training.

\begin{figure*}[t]
\centering
\includegraphics[width=\textwidth]{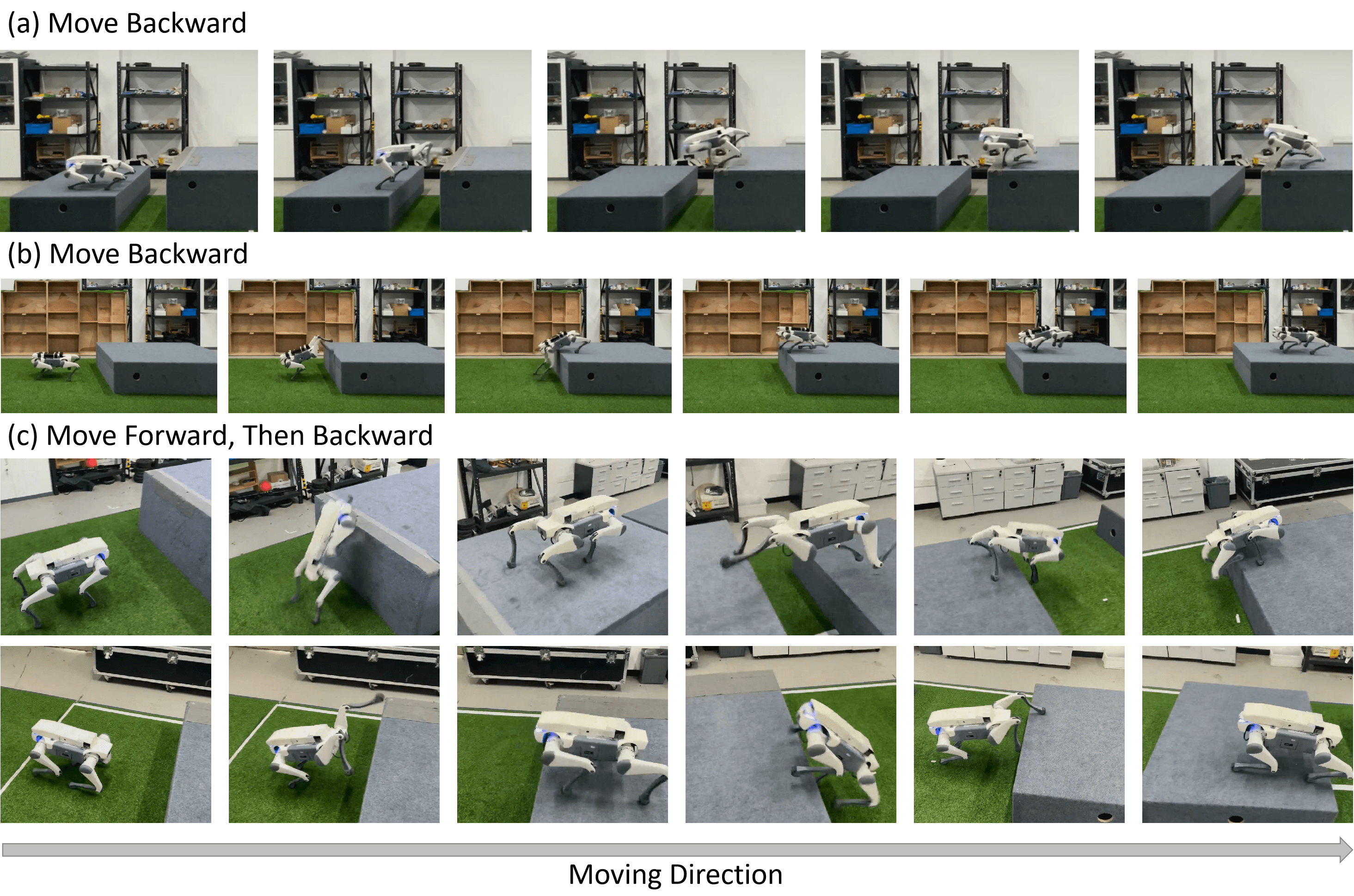}
\vspace{-0.5cm}
\captionsetup{font=footnotesize}
\caption{Real-world deployment on a robot equipped with a forward-facing depth camera. For clarity, all sequences are visualized such that the robot moves from left to right.}
\vspace{-0.5cm}
\label{fig_4}
\end{figure*}

\subsection{Real-World Deployment}

We deploy the controller on a DEEP Robotics Lite3 low-cost quadruped equipped with an onboard RK3588 processor and a forward-facing RealSense depth camera. Images are streamed at \(10\,\mathrm{Hz}\) and the control loop runs at \(50\,\mathrm{Hz}\). The policy executes strictly recurrent inference with a fixed-size memory state and constant per-step computation, which keeps the hardware control loop stable.

As shown in Fig.~\ref{fig_4}, we evaluate on multiple terrains, including a \(0.6\,\mathrm{m}\) high step, a \(0.6\,\mathrm{m}\) wide gap, and composite layouts that combine steps and gaps. The longest trails are close to \(8\,\mathrm{m}\), which is nearly the limit of our test area. In Fig.~\ref{fig_4}(a), the robot jumps backward across a gap with a \(0.3\,\mathrm{m}\) height drop between two edges, which is rarely demonstrated even in forward perceptive locomotion by prior works. Our system not only performs precise forward jumps with perception, but also executes the same maneuver in reverse by relying on accurate memory. In Fig.~\ref{fig_4}(b), the robot jumps backward onto a high step. In Fig.~\ref{fig_4}(c), we show the full pipeline of first moving forward to memorize the surroundings and then moving backward, climbing multiple obstacles while retrieving the correct information from the memory state. Despite cumulative real-world noise, the memory module remains robust and adaptive.

\section{CONCLUSION}

In this work, we present Look Forward to Walk Backward (LF2WB), a single-stage framework that uses only forward vision to build and maintain an efficient terrain memory and retrieve it for collision-free backward locomotion without rearward sensing. A delta-rule selective update keeps this memory compact, while chunkwise-parallel training makes learning efficient. Serial recurrent inference with constant-size states makes the method well-suited for efficient deployment on onboard processors. In our experiments, LF2WB outperforms other recurrent networks and Transformer-XL under varied forward/backward horizon stresses. We further demonstrate zero-shot deployment on a DEEP Robotics Lite3 quadruped robot, which relies solely on a forward depth camera and the learned memory to execute agile backward maneuvers over challenging composite terrains that combine high steps and wide gaps.

However, certain limitations still exist. The memory is implicit, so its effective horizon is not formally guaranteed. We sometimes observe surprisingly strong generalization. For example, although we train only forward and backward motion, the robot still shows some memory capability when turning. But we also see occasional failures on seemingly simple cases. We believe broader task diversity and scene coverage during training will improve reliability. Incorporating estimator uncertainty into the concurrent training loop could further guide the actor to respect the estimator’s capability limits and enhance overall robustness.


\bibliographystyle{IEEEtran}

\bibliography{references}

\end{document}